\documentclass[letterpaper, 10 pt, conference]{ieeeconf}

\IEEEoverridecommandlockouts                              

\usepackage{amsmath,amsfonts}
\usepackage{algorithmic}
\usepackage{algorithm}
\usepackage{array}
\usepackage{textcomp}
\usepackage{stfloats}
\usepackage{url}
\usepackage{verbatim}
\usepackage{graphicx}
\usepackage{cite}

\usepackage{amsmath}
\usepackage{amssymb}
\usepackage[colorinlistoftodos]{todonotes}
\usepackage{hyperref}
\usepackage{multirow}
\usepackage{booktabs}
\usepackage[colorinlistoftodos]{todonotes}
\usepackage{siunitx}
\usepackage{url}
\usepackage{import,bm}
\usepackage[super]{nth}
\usepackage{romannum}
\usepackage{accents}
\usepackage{verbatim}
\usepackage{tikz}
\usetikzlibrary{automata, positioning, arrows, shapes.geometric, arrows.meta, positioning, calc, intersections, spath3}
\usepackage{tkz-euclide}
\usepackage{subcaption}
\usepackage{graphicx}
\usepackage{cite}
\usepackage{makecell}
\usepackage{caption}
\usepackage{ragged2e}
\usepackage{soul}

\newcommand{\mt}[1]{\bm{#1}}

\AtBeginShipoutNext{\AtBeginShipoutUpperLeft{%
  \put(\dimexpr\paperwidth-1cm\relax,-1.5cm){\makebox[0pt][r]{\framebox{Accepted for publication at the IEEE International Conference on Robotics and Automation (ICRA), 2024.}}}%
}}

\begin{document}

\title{Exoskeleton-Mediated Physical Human-Human Interaction for a Sit-to-Stand Rehabilitation Task}

\author{Lorenzo Vianello$^1$, Emek Barış Küçüktabak$^{1,2}$, Matthew Short$^{1,3}$, Clément Lhoste$^{1}$, Lorenzo Amato$^{1,4}$,\\ Kevin~Lynch$^{2}$, and Jose Pons$^{1,2,3}$
\thanks{$^1$ Legs and Walking Lab of Shirley Ryan AbilityLab, Chicago, IL, USA}
\thanks{$^2$ Center for Robotics and Biosystems of Northwestern University, Evanston, IL, USA}
\thanks{$^3$ Department of Biomedical Engineering, Northwestern University, Evanston, IL, USA}
\thanks{$^4$ The Biorobotics Institute, Scuola Superiore Sant’Anna, 56025 Pontedera, Italy \textit{and} Department of Excellence in Robotics \& AI, Scuola Superiore Sant'Anna, 56127 Pisa, Italy}
}

\markboth{Journal of \LaTeX\ Class Files,~Vol.~14, No.~8, August~2021}%
{Shell \MakeLowercase{\textit{et al.}}: A Sample Article Using IEEEtran.cls for IEEE Journals}


\maketitle

\begin{abstract}
Sit-to-Stand (StS) is a fundamental daily activity that can be challenging for stroke survivors due to strength, motor control, and proprioception deficits in their lower limbs. Existing therapies involve repetitive StS exercises, but these can be physically demanding for therapists while assistive devices may limit patient participation and hinder motor learning. To address these challenges, this work proposes the use of two lower-limb exoskeletons to mediate physical interaction between therapists and patients during a StS rehabilitative task. This approach offers several advantages, including improved therapist-patient interaction, safety enforcement, and performance quantification. The whole body control of the two exoskeletons transmits online feedback between the two users, but at the same time assists in movement and ensures balance, and thus helping subjects with greater difficulty. In this study we present the architecture of the framework, presenting and discussing some technical choices made in the design.
\end{abstract}

\section{Introduction}
Transitioning from sitting to standing is fundamental to living independently and is one of the most frequently executed functional tasks. This task, commonly referred to as Sit-to-Stand (StS), can be difficult to perform for post-stroke individuals due to deficits in strength, motor control and proprioception of the lower limb. Compared to healthy individuals, post-stroke patients present many problems related to their StS ability: asymmetric force and weight distribution, reduced peak vertical reaction force, increased time to complete the movement, and a larger mediolateral center of pressure displacement~\cite{boukadida2015determinants}.
In addition, there is a correlation between the ability to perform StS tasks and clinical balance scales in post-stroke patients ~\cite{cheng1998sit}.


Typically, the therapist's role in StS tasks is to instruct and guide the patient toward more correct movements via physical assistance and helps those with greater disabilities by providing partial to full weight support. These approaches require a large amount of physical effort from the therapist which can, over time, lead to work-related injuries~\cite{mccrory2014work}. 

For this reason, assistive devices are usually offered to patients who lack the strength to support their own weight~\cite{burnfield2013comparative}. Nevertheless, many of these devices require the patient to play mostly a passive role, resulting in a smaller activation of motor learning mechanisms~\cite{fray2019evaluation}.
Moreover, they usually do not allow the therapist to provide external physical guidance or assistance. Eliminating the physical patient-therapist interaction significantly limits the usage of therapist's knowledge and ability, and therefore the efficacy of the rehabilitation.

\begin{figure}[t]
    \centering
    \hspace{-0.3cm}
    \includegraphics[width=0.48\textwidth]{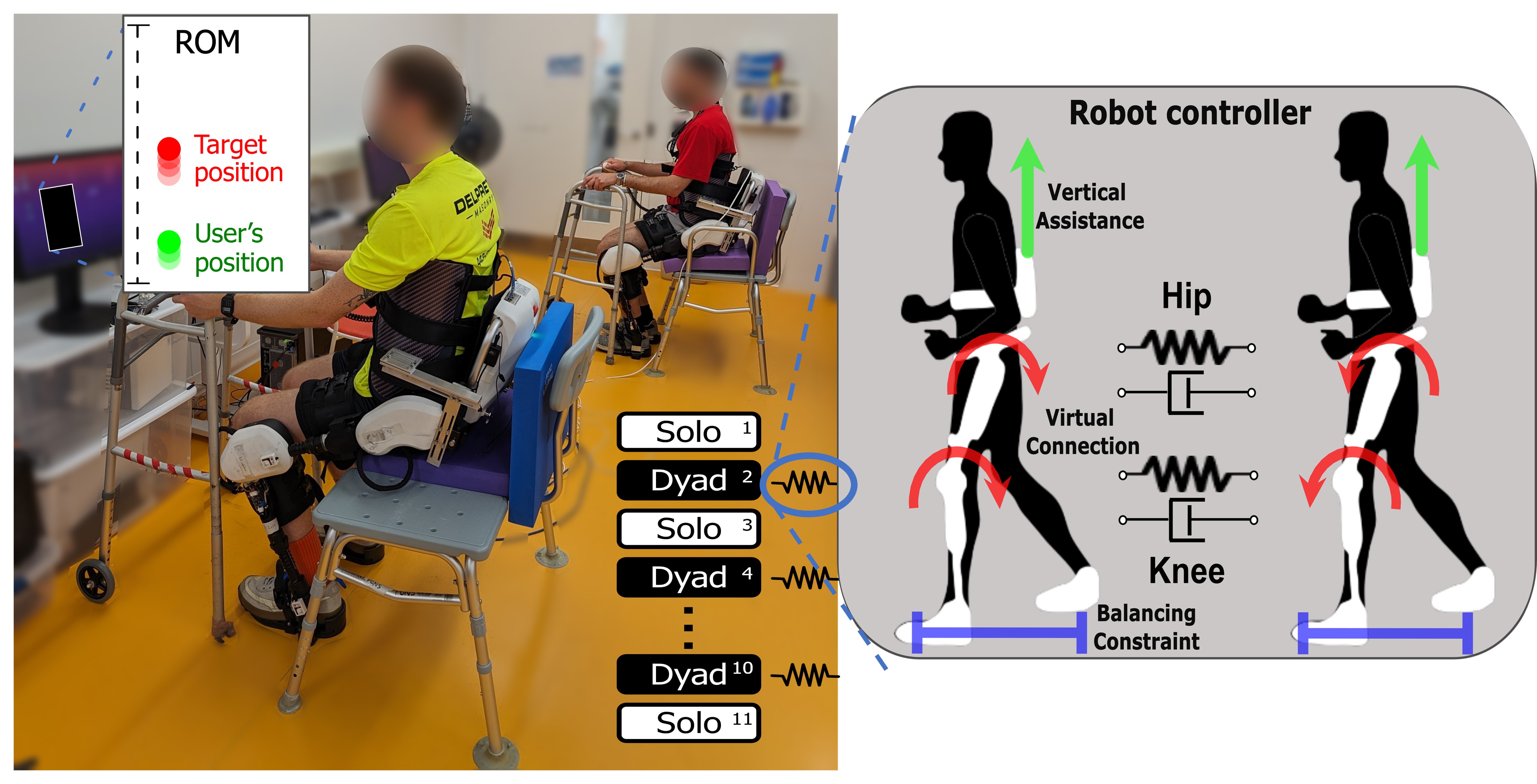}
    \caption{Sit-to-Stand rehabilitative framework: two exoskeletons are used to mediate the interaction between two users performing a tracking task in which the position of the two Center of Masses is used as a reference. The controller of the two exoskeletons produces a threefold function: 1) they virtually connect the movements of the two users; 2) they produce vertical assistance on the two by limiting fatigue; and 3) they preserve balancing.  }
    \label{fig:conceptual_image}
    \vspace{-0.2cm}
\end{figure}

An infrastructure with two exoskeletons can be used to mediate physical interaction between a patient and therapist similar to the previously proposed human-robot-human systems~\cite{Kucuktabak2021}. Such an infrastructure has several advantages. First, it allows a therapist to feel and guide their patient on multiple joints and contact points simultaneously. Second, the exoskeleton can be programmed to further assist and enforce defined safety criteria which are particularly relevant during StS tasks. Third, interfacing with exoskeletons facilitates the objective quantification of motor performance over many repetitions, increasing reproducibility. Last but not least, this infrastructure could enhance tele-rehabilitation practices, allowing therapists and patients to physically interact in different remote settings (e.g., at home, in the clinic).

To the best of our knowledge, there hasn't been any prior presentation of a rehabilitation framework using overground lower-limb exoskeletons to mediate physical interaction between therapists and patients for StS task. In fact, while overground systems offer the benefit of promoting natural movements while engaging the balance, they can be challenging to control~\cite{camardella2021gait, sharifi2021adaptive, andrade2021trajectory}.
The whole body controller of the exoskeleton must not only be designed to consider the degree of impairment of the patient, but at the same time to meet safety prerequisites such as maintaining balance~\cite{huo2021impedance}.

In our prior research, we introduced a control algorithm that controls the forces between a floating-base lower-limb exoskeleton and its wearer while taking into account actuation constraints and safety requirements such as velocity and torque limits~\cite{kuccuktabak2023haptic}. 
Building on this formalization, we introduced the practical application of exoskeleton-mediated physical interaction between two users~\cite{kuccuktabak2023virtual}. This approach combines the benefits of a whole-body controller, including constraint and force regulation, with spring-damper rendering to couple the joints of two users. Exoskeleton-mediated physical interaction aims to explore how task performance and motor learning effects, initially assessed on simpler robots with just one joint, can be scaled to address more complex, bilateral tasks~\cite{Kucuktabak2021}. Indeed, robot-mediated physical interaction is acknowledged for its potential to facilitate knowledge transfer between individuals while also enabling the recording of movements and the repetition of exercises~\cite{short2023haptic, kim2022effect}. 

In this context, we propose a StS framework for physical rehabilitation that virtually connects two lower-limb exoskeletons operated by human users. To promote repetitive training, we designed a task in which users control the vertical position of their Center of Mass (CoM) with the help of continuous visual feedback. In this work, we compared two haptic interaction strategies: (1) virtual connections applied directly between the joints of the two participants and (2) virtual connections linking the vertical position of the two CoMs. The haptic interaction is represented as a spring-damper system, and several compliance profiles are tested. The proposed framework can scale the assistance provided to patients according to their functional level and uses a balancing recovery strategy 
We present the structure of our framework and the results of this preliminary testing.

\section{METHODS}

\subsection{Exoskeleton Whole Body Model}
For the StS task, we considered the exoskeleton  in a state of double stance. The equation of motion of the exoskeleton during this state is described as:
\begin{equation}
     \mt{M}_\text{ds}(\mt{q})\ddot{\mt{q}} + \mt{b}_\text{ds}(\mt{q},\dot{\mt{q}}) + \mt{g}_\text{ds}(\mt{q}, \mt{\alpha})  = S^T(\mt{\tau}_\text{motor} + \mt{\tau}_\text{friction}) +  \mt{\tau}_\text{int}.
    \label{eq:FlyingEoMFriction}
\end{equation}
where $\mt{q} = [\theta_0, \theta_1, \theta_2, \theta_3, \theta_4]$ are the generalized coordinates corresponding to the backpack, hip, and knee angles. The matrix and vectors $\mt{M}_{ds} \in \mathbb{R}^{5\times 5}, \mt{b}_{ds} \in \mathbb{R}^5, \mt{g}_{ds} \in \mathbb{R}^5$, are the mass matrix,  Coriolis and centrifugal torques, and gravitational torques respectively during double stance. The variable $\mt{\tau}_\text{motor} \in \mathbb{R}^4$ corresponds to the motor torques sent to the driver, $\mt{\tau}_\text{friction} \in \mathbb{R}^4$ is the friction component of the torque, $S = [0_{4\times 1}, \mathbb{I}_{4\times 4}]$ is a selection matrix of actuated joints, and $\mt{\tau}_\text{int} \in \mathbb{R}^5$ is a vector of interaction torques applied to the exoskeleton by the user.

The parameter $\mt{\alpha}$ is used to quantify the amount of vertical assistance the exoskeleton provides to the user expressed in $kg$.
This parameter is used to delay or avoid the onset of fatigue in StS exercise. In fact, fatigue may impair motor learning during rehabilitation. This parameter can be easily scaled during rehabilitation by increasing or decreasing assistance depending on the patient's level of impairment.

\subsection{Human-Exoskeleton Interaction}
The model presented in Eq.~\eqref{eq:FlyingEoMFriction} is used to control the joint acceleration such that it follows a desired interaction profile. With this purpose, we implemented the virtual mass controller presented in~\cite{kuccuktabak2023haptic} to simulate a
desired virtual mass or inertia on the joints of the exoskeleton:
\begin{equation}
    \ddot{\mt{q}}^* = \mt{M}_\text{virt}^{-1}(\mt{\tau}_\text{int} - \mt{\tau}_\text{int}^*),
    \label{eq:desired_acc}
\end{equation}
where $\mt{\tau}_\text{int}^*$ is the desired interaction torque. 
For instance, whenever we want to minimize the interaction between the exoskeleton and the user, we apply zero desired interaction torque: 
\begin{equation} \label{eq:transparent}
        \mt{\tau}_\text{int}^* =0.
\end{equation}
This condition is typically referred to as transparent control; in this paper, it is used as a baseline for motor performance.

In many rehabilitation exercises, the exoskeleton is commanded to interact with the user in a certain way (e.g., providing assistance along a trajectory \cite{medrano2023real}). In our proposed framework of virtual interaction between two exoskeletons ($A$, $B$), the desired interaction with the user is defined as a virtual spring-damper between the configuration of the two exoskeletons. This behavior can act directly at joint level by defining:
\begin{equation} \label{eq:spring_collab}
    \begin{split}
        \mt{\tau}^\text{int}_A &= K_\text{virt,q}(\theta_B - \theta_A) + C_\text{virt,q}(\dot\theta_B - \dot\theta_A) \\
        \mt{\tau}^\text{int}_B &= -\mt{\tau}^\text{int}_A,
    \end{split}
\end{equation}
where $\theta_{A,B}, \dot\theta_{A, B}$ are respectively the joints angle and velocities of the robots $A$ and $B$. The variables $ K_\text{virt,q}$ and $C_\text{virt,q}$ are the constants of the rendered spring and damper between the users, respectively.

In addition to the joint space interaction, it is also possible to design the interaction in the task space. In other words, virtual elements can be rendered between the two points of the exoskeletons in the sagittal plane. In this work, we rendered spring and damper between the vertical positions ($z$) of their CoM.

\begin{equation} \label{eq:spring_collab_task}
    \begin{split}
        \mt{F}^\text{int}_A &=  K_\text{virt,z}(z_{B} - z_{A}) + C_\text{virt,z}(\dot z_{B} - \dot z_{A})  \\
        \mt{F}^\text{int}_B &= -\mt{F}^\text{int}_A \\
        \mt{\tau}^\text{int}_A &= \mathbb{J}_{\text{CoM}}^z(\mt{\theta}_A)^T \mt{F}^\text{int}_A \\ \mt{\tau}^\text{int}_B &= \mathbb{J}_{\text{CoM}}^z(\mt{\theta}_B)^T \mt{F}^\text{int}_B,
    \end{split}
\end{equation}
where $z_{A,B}, \dot{z}_{A, B}$ are respectively the vertical positions and velocities of the robots $A$ and $B$ and $\mathbb{J}_{\text{CoM}}^z(.) \in \mathbb{R}^{1\times5}$ is the vertical component of CoM Jacobian. This type of interaction allows freedom of movement in the null space of the vertical component of motion and a larger selection of the postures that can meet the given position of the CoM.  

\subsection{Quadratic Optimal Controller}
To track the desired generalized acceleration presented in Eq.~\eqref{eq:desired_acc} under physical and safety constraints, we use real-time constrained optimization with the OSQP library~\cite{osqp} to solve for motor torques and joint accelerations, $\textbf{x} = (\ddot{\mt{q}}, \mt{\tau}_{\text{motor}}) \in \mathbb{R}^9$:
\begin{equation}
\min_\textbf{x} \lVert f(\textbf{x})\rVert \qquad \text{ s.t. } \textbf{A} \textbf{x} = \textbf{b} \textit{ and } \textbf{D} \textbf{x}  \leq \textbf{f}.
\label{eq:const_opt}
\end{equation}
The objective function $\lVert f(\textbf{x}) \rVert$ is the 2-norm of the difference between the actual and desired generalized accelerations,
\begin{equation}
f(\textbf{x}) =  [\mathbb{I}_{5\times5},\; 0_{5\times4}]\textbf{x} - \ddot{\mt{q}}^*.
\label{eq:obj_function}
\end{equation}

The equality constraint ensures that optimized variables satisfy the equation of motion (Eq.~\eqref{eq:FlyingEoMFriction}), 
\begin{equation} \label{eq:task_EOM}
\begin{split}
    \textbf{A} = \textbf{A}_{\text{EOM}} &= [\mt{M}_\text{ds},\: -S^\top], \\
    \textbf{b} =\textbf{b}_{\text{EOM}} &= [-\mt{b}_\text{ds} -\mt{g}_\text{ds} + \mt{\tau}_{\text{int}} + S^\top\mt{\tau}_\text{friction}].
\end{split}
\end{equation}

Inequality constraints arise from physical and safety constraints on motor torque, power and velocity as presented in our previous work~\cite{kuccuktabak2023haptic}.
In this paper we extend our previous formulation including a constraint on the joints positions that avoids reaching the robot's mechanical joint limits and creating unexpected interactions. This is achieved by constraining the joint accelerations such that the joint position limits will not be violated in $\beta \Delta t$ amount of time, where $\Delta t$ is the control period.
\begin{equation}
\begin{split}
    \textbf{D}_{{\mt{q}}^+} = [\mathbb{I}_{5\times5},\: 0_{5\times4}], \; \textbf{f}_{{\mt{q}}^+} &= 2 \frac{{\mt{q}}_{\text{max}} - {\mt{q}} - \dot{\mt{q}}\beta\Delta t }{\beta^2\Delta t^2} \\ \textbf{D}_{{\mt{q}}^-} = [-\mathbb{I}_{5\times5},\: 0_{5\times4}], \; \textbf{f}_{{\mt{q}}^-} &= -2 \frac{{\mt{q}}_{\text{min}} - {\mt{q}} - \dot{\mt{q}}\beta\Delta t }{\beta^2\Delta t^2}, 
    \label{eq:task_pos}
\end{split}
\end{equation}

We also added a safety constraint on the Cartesian position of the CoM that ensures balancing in the sagittal plane which is presented in the following subsection.

\begin{figure}[t]
    \vspace{0.3cm}
    \centering
    \includegraphics[width =0.99\linewidth]{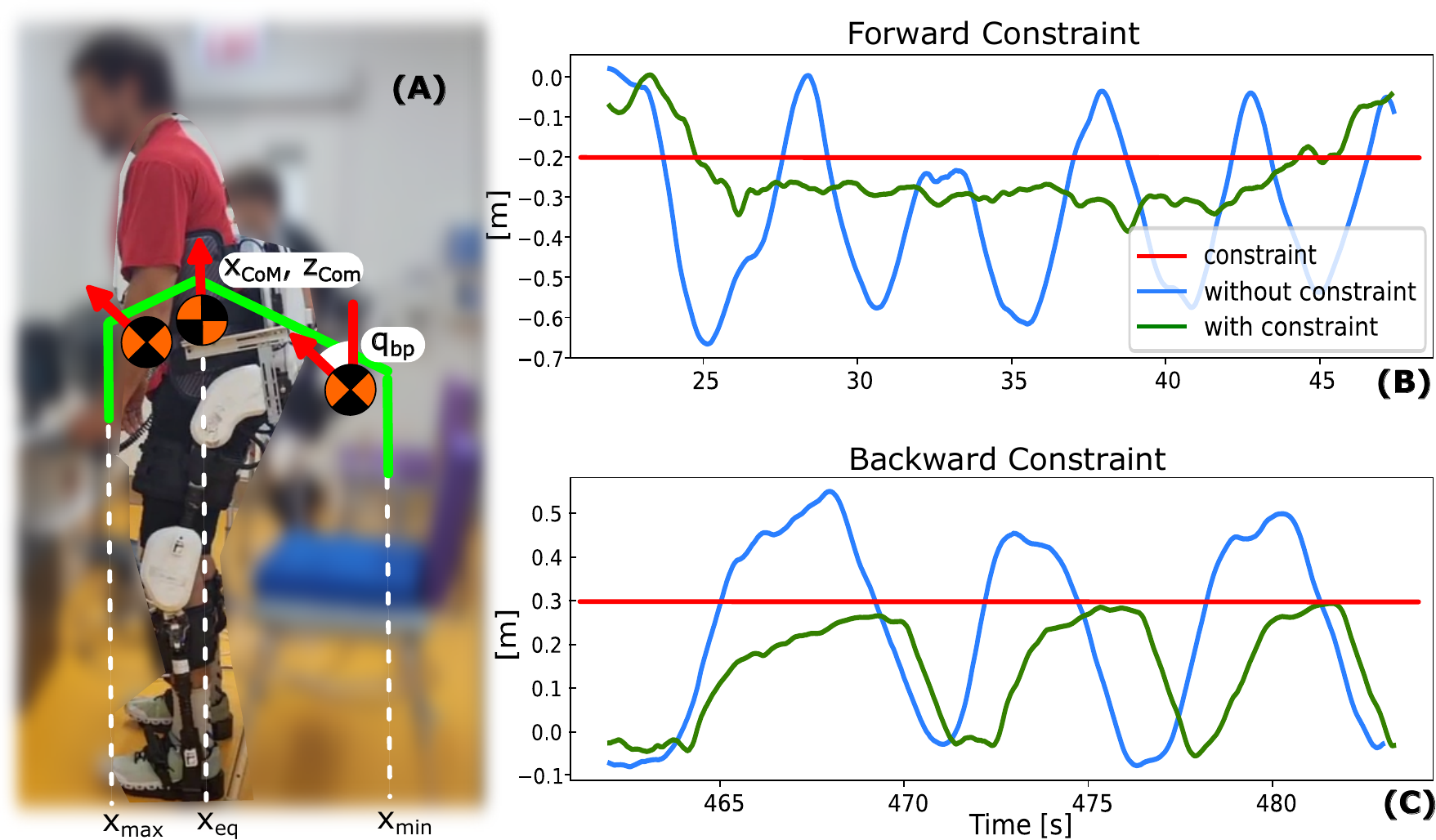}
    \caption{Balancing constraint for the StS task: Fig. (A) shows how the pose (position and orientation) of the CoM is constrained in both the horizontal and vertical planes; Fig. (B) shows how the position is constrained in the forward movement to avoid falling; Fig. (C) shows how the backward movement is constrained but still leaves the possibility of performing the StS task. }
    \label{fig:balancing_constraint}
\end{figure}

\subsection{Balancing Constraint}
\label{sec:balancing}

Human-exoskeleton balancing is a critical aspect of exoskeleton control to provide seamless coordination between the wearer and the exoskeleton for enhanced stability and mobility~\cite{hamza2020balance}. The main solutions proposed are based on fall recognition and balance recovery and satisfaction~\cite{farkhatdinov2019assisting}.
Many of these algorithms in the literature are based on strategies already used to control humanoid robots~\cite{ramos2019dynamic, penco2018robust, vianello2021human}. In fact, balancing is usually satisfied by constraining the position of the CoM within a support region defined by the position of the feet and the activity being performed~\cite{harib2018feedback,xu2017adaptive, deng2018human, menga2018lower}.  
Many of the algorithms presented for applications in rehabilitation give little freedom of motion to the patient and tend to prefer positional control strategies~\cite{rodriguez2022wearable}. This, on the one hand, allows greater safety for the wearer, but on the other hand limits their contribution by enforcing passive repetitions of a movement. 
In this scenario, Huo et al.~\cite{huo2021impedance} presented a system that prioritizes the wearer's control and delivers adequate assistance for power and balance. 

In this work, we employ a similar approach  in order to satisfy the balancing of the exoskeleton-human couple. We constrain the motion of the Divergent Component of Motion (DCM) to remain inside the support region of the sagittal plane of the exoskeleton. Since all the actuation of the exoskeleton is on the sagittal plane, we implemented constraints only on that plane. Similarly to~\cite{ishiguro2020bilateral} we define the DCM as 
\begin{equation}
    \mt{p}_{\text{DCM}} = \mt{p}_{\text{CoM}} + (\sqrt{h/g}) \dot{\mt{p}}_{\text{CoM}}
\end{equation}
where $\mt{p}_{\text{CoM}} = (x_{\text{CoM}}, z_{\text{CoM}})$ is the position of the CoM in the sagittal plane (Fig. \ref{fig:balancing_constraint}A) while $h$ is the height of the linear inverted pendulum which approximates the exoskeleton-human couple and is assumed as the height of the CoM of the exoskeleton as the human is at an upright posture.
This aspect characterizes the pendulum's susceptibility to losing balance, with a larger $h$ indicating greater instability. In the case of patients, this parameter may be increased, increasing the demand for assistance, especially for those facing greater difficulties.

To constraint the position of the DCM we implemented the following inequality constraint ($Dx \leq f$):
\begin{equation}
\begin{split}
    \textbf{D}_{\text{CoM}^+} &= [\mathbb{J}_{\text{CoM}},\: 0_{2\times4}], \qquad \textbf{D}_{\text{CoM}^-} = - \textbf{D}_{\text{CoM}^+},\\
    \textbf{f}_{\text{DCM}^+} &= 2 \frac{\mt{p}^+ -\mt{p}_{\text{DCM}}  - \dot{\mt{p}}_\text{DCM} \beta\Delta t}{\beta^2\Delta t^2} - \dot{\mathbb{J}}_\text{CoM}\dot{\mt{q}}, \\  
     \textbf{f}_{\text{DCM}^-} &= -2 \frac{\mt{p}^- -\mt{p}_{\text{DCM}}  - \dot{\mt{p}}_\text{DCM} \beta\Delta t}{\beta^2\Delta t^2} + \dot{\mathbb{J}}_\text{CoM}\dot{\mt{q}}, \\
\end{split}
\end{equation}

\begin{equation*}
    \textbf{f}_{\text{DCM}^-} \leq \mathbb{J}_{\text{CoM}} \ddot{\mt{q}} \leq \textbf{f}_{\text{DCM}^+}
\end{equation*}
where $\mt{p}^+, \mt{p}^-$ are the maximum and minimum limits for the DCM position, respectively. This formalization is similar to the joint position limit expressed in Eq.~\eqref{eq:task_pos}. The joint accelerations are constrained such that the CoM pose does not exceed the defined limits in $\beta \Delta t$ amount of time.

Horizontal DCM position is defined as constant along the horizontal ($x$) axes while the vertical component ($z$) is defined as a function of the horizontal position. Similarly, to limit the over extension of the backpack angle we made the upper limit of the backpack angle ($\mt{q}_{bp}^+$) variable from the horizontal position. Specifically, the two constraints are defined as linear functions that with a maximum at the maximum equilibrium position of the user ($x_{eq}$) and two minima in the limits of the horizontal position of the sagittal plane. This is formalized as follows and visualized in Fig.~\ref{fig:balancing_constraint}A.
\begin{equation}
\begin{split}
    \mt{p}_{z}^+ &= z_{max} - ||a_z (x -x_{eq})||,\\  
    \mt{q}_{bp}^+ &= \theta_{max} - ||a_{\theta} (x -x_{eq})||,\\
\end{split}
\end{equation}

We defined the constraint in that way to compensate for the robot's uncontrollable movements (non-actuated  ankle joint). A more in-depth discussion of the implications of such a constraint is covered in Sec. \ref{sec:results}.

\begin{figure}[t]
    \centering
    \vspace{0.2cm}
    \includegraphics[width= 0.85\linewidth]{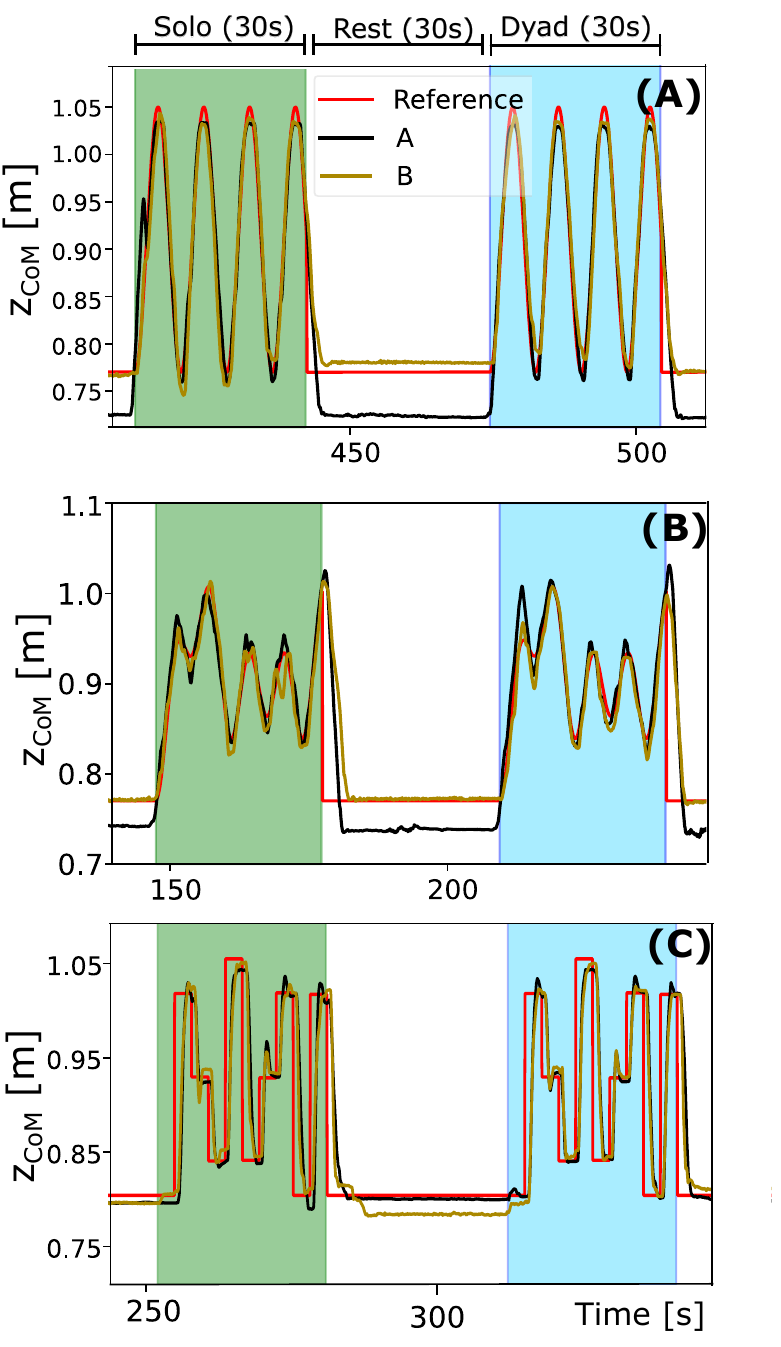}
    \caption{Composition of trials with different visual feedback profiles: each trial consists of an equal time (30 s) of performing the motion alone, resting, and performing the motion while subjected to virtual interaction. Three types of reference were tested: (A) simple sine, (B) composite sine, and (C) discrete tasks in the range of motion.}
    \label{fig:trials}
\end{figure}

\begin{table}
\begin{center}
\vspace{0.2cm}
\caption{Values of the gains used in the experimentation.}\vspace{1ex}
    \label{tab:controller_gains}
\begin{tabular}{c|c|c|c}
\cline{1-4}
\rule{0pt}{2ex}  
Parameter & Value & Parameter & Value    \\ \cline{1-4}
$\mt{K}_\text{virt,q}^\text{soft}$ & $ \mathbb{I}_{4\times4} \times 30 [\frac{\text{Nm}}{\text{rad}}]$ & $\mt{C}_\text{virt,q}^\text{soft}$ &  $\mathbb{I}_{4\times4} \times 4 [\frac{\text{Nms}}{\text{rad}}]$    \\
$\mt{K}_\text{virt,q}^\text{stiff}$ &  $ \mathbb{I}_{4\times4} \times 70 [\frac{\text{Nm}}{\text{rad}}]$   & $\mt{C}_\text{virt,q}^\text{stiff}$ &  $ \mathbb{I}_{4\times4} \times 10  [\frac{\text{Nms}}{\text{rad}}]$     \\
${K}_\text{virt,z}$ & 1000  $[\frac{\text{N}}{\text{m}}]$  & ${C}_\text{virt,z}$ & 50   $[\frac{\text{Ns}}{\text{m}}]$    \\
$p^+$  & 0.2 [m] &$\mt{p}^-$ & 0.3 [m] \\ \cline{1-4}
\end{tabular}
\end{center}
\vspace{-0.5cm}
\end{table}

\subsection{Experimental Protocol}

In this paper, we performed exploratory work to define a StS exercise using exoskeletons to mediate patient-therapist interaction. Two users wore lower-limb exoskeletons (X2, Fourier Intelligence, Singapore). The exoskeleton has 4 actuated Degrees of Freedom (DoF): the two hips and the knees. The physical joint limits are set at $120^{\circ}$ and $-40^{\circ}$ for the hip joint and $0^{\circ}$ and $-120^{\circ}$ for the knee joint.  The presented controller was implemented on a ROS and C++ based software stack named CANOpen Robot Controller~(CORC)~\cite{FongCanOpenDevelopment}. The gains used in the controller are presented in Table~\ref{tab:controller_gains}.

The distance between the chair and the ankle was adjusted according to the participant's preference, while the height of the chair was chosen according to knee height. Dyads were seated side by side, facing two monitors which displayed visual feedback for the tracking task (Fig.~\ref{fig:conceptual_image}). To validate the usability of the exoskeleton by therapists, users wore the exoskeleton without external help. At this point, users familiarized themselves for two minutes with the StS task wearing the exoskeleton. During this time, the range of motion was recorded for visual reference and to define balancing constraints.

\begin{figure*}[t]
    \centering
    \vspace{0.2cm}
    \includegraphics[width = 0.9\linewidth]{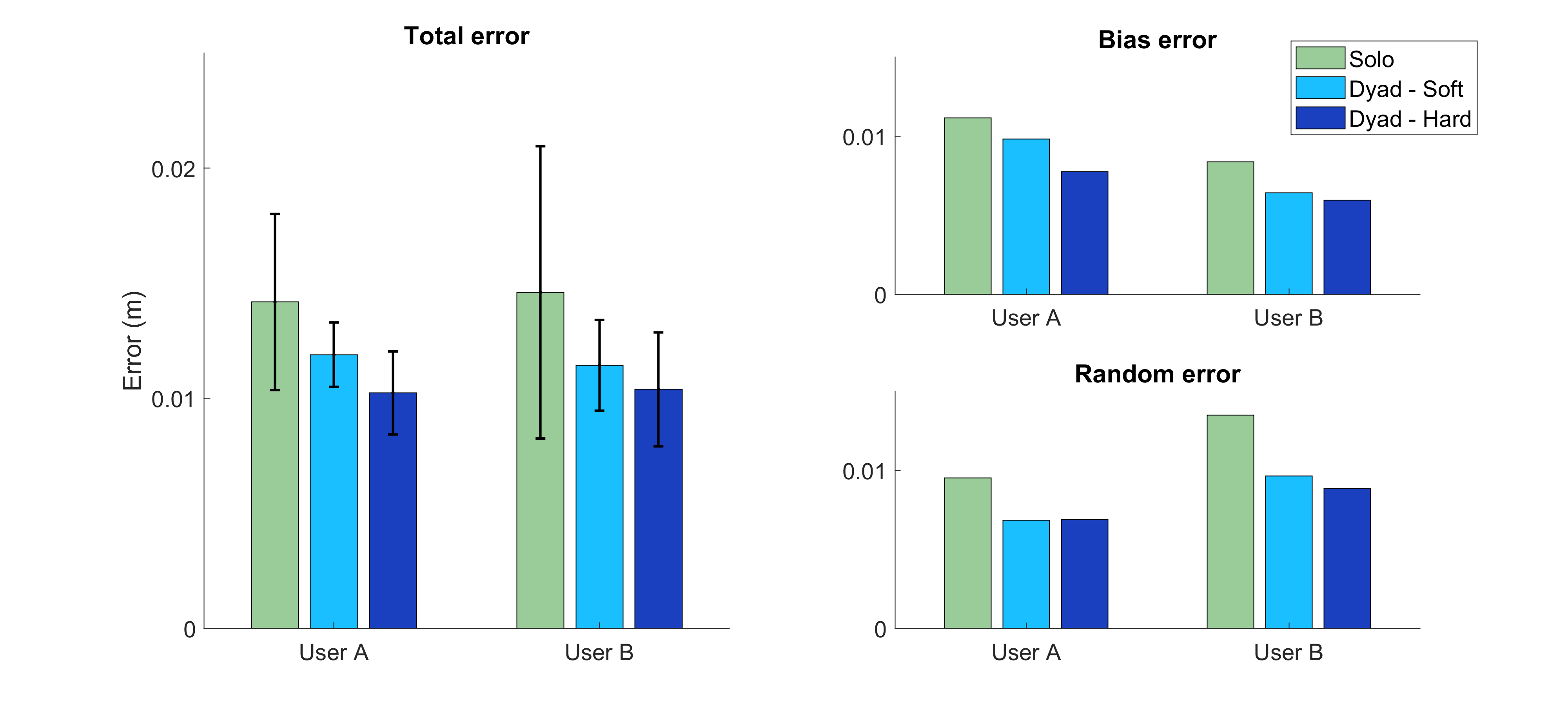}
    \caption{Tracking performance improved when two pilot users were coupled via joint space connection compared to no connection. The left panel shows the total error across repetitions of the StS exercise (mean $\pm$ SD) for two users during solo and dyad trials under two connection stiffnesses. Right panels show the error separated into bias and random across conditions.}
    \label{fig:RMSE}
    \vspace{-0.5cm}
\end{figure*}

Each trial consisted of the alternating succession of 3 periods lasting 30 seconds (Fig.~\ref{fig:trials}): one period in which users performed the exercise alone (transparent exoskeleton, $\mt{\tau}^\text{int}=0$), one period in which the exoskeleton is used to virtually connect the two users (Eq.~\eqref{eq:spring_collab},~\eqref{eq:spring_collab_task}), and a rest period. Each experimental condition is composed by 8 trials with 30 seconds of rest between each trial in which the two users performed the task in parallel. 
During the tracking task both users were provided with visual feedback of the target vertical CoM position (red point) and their own instantaneous vertical CoM position (green dot) as shown in Fig.~\ref{fig:conceptual_image}. Participants were instructed to match their CoM position to the target position as accurately as possible while the target moved along a vertical bar. The upper and lower bounds of the target movement were determined by the range of motion (ROM) of the two participants. Specifically, we used the distance between the maximum of the minima and the minimum of maxima to define a common ROM. We chose to use this convention to model the therapist-patient case, in which the therapist is more likely to adapt to the patient's joint limits. 
Furthermore, in this work, we have chosen not to use a mapping that amplifies one's movements over the other to avoid discontinuity with the perceived haptic feedback. However, in future research, we intend to investigate how to identify transfer function between the movements (and haptic feedback) of a patient and a therapist interacting in this manner. Three types of references were tested with the common ROM as the amplitude: (1) a simple sine function, (2) multi-sine function with different frequencies, and (3) a step function with discrete values in the ROM, 
\begin{equation}
\begin{split}
    z_{\text{des}} = \text{ROM} \times[&\text{sin}(0.5(2\pi)(t+t_{r})+\phi_{1})] ,\\
    z_{\text{des}} = \text{ROM}\times[&0.33\text{sin}(0.50(2\pi)t) \\
    +&0.33\text{sin}(0.20(2\pi)t)\\
    +&0.33\text{sin}(0.16(2\pi)t)] ,\\
\end{split}
\label{eq:multisine}
\end{equation}
where $z_\text{des}$ is the target vertical CoM position at a given time. 

In this preliminary study, we compared different configurations to define the usability of the StS infrastructure. With this intent, we tested different interaction profiles (i.e., stiffness and damping) as shown in Tab. \ref{tab:controller_gains}.
We present the results and discuss measures of performance which are relevant for a StS exercise. In particular, we are interested in a user's tracking errors while they attempt to accurately follow the visually-displayed target during the exercise. We use the root-mean-square of the difference between each individual's actual trajectory and the target trajectory to represent the total error during each tracking trial. In addition, we distinguish between the type of error, bias (i.e., systematic) and random, as they each contribute to the total tracking error. Bias error gives an indication of repeated mistakes by the user (e.g., undershooting the peak target position) across tracking trials. Random error indicates the consistency of the movement across trials, as low random error would suggest the user smoothly and repeatably pursues the target from trial to trial. We present these three categories of error (i.e., total, bias and random) for each of the stiffness configurations during the joint space interaction to show how haptic coupling influences tracking performance for an example dyad.

\section{RESULTS}
\label{sec:results}

Comparing performances with and without haptic interaction during the tracking task, our preliminary results are consistent with previous work in dyadic tracking at the ankle~\cite{short2023haptic}. In fact, not only does the interaction seem to improve one's ability to follow a reference, but also the degree of improvement appears related to the magnitude of virtual stiffness of the connection. It is not the intent of this paper to validate these findings for lower-limb exoskeletons but we report here a change in user performance for one dyad (Fig.~\ref{fig:RMSE}).

\begin{figure}[t]
  \centering
  \vspace{0.2cm}
  \includegraphics[width=.75\linewidth]{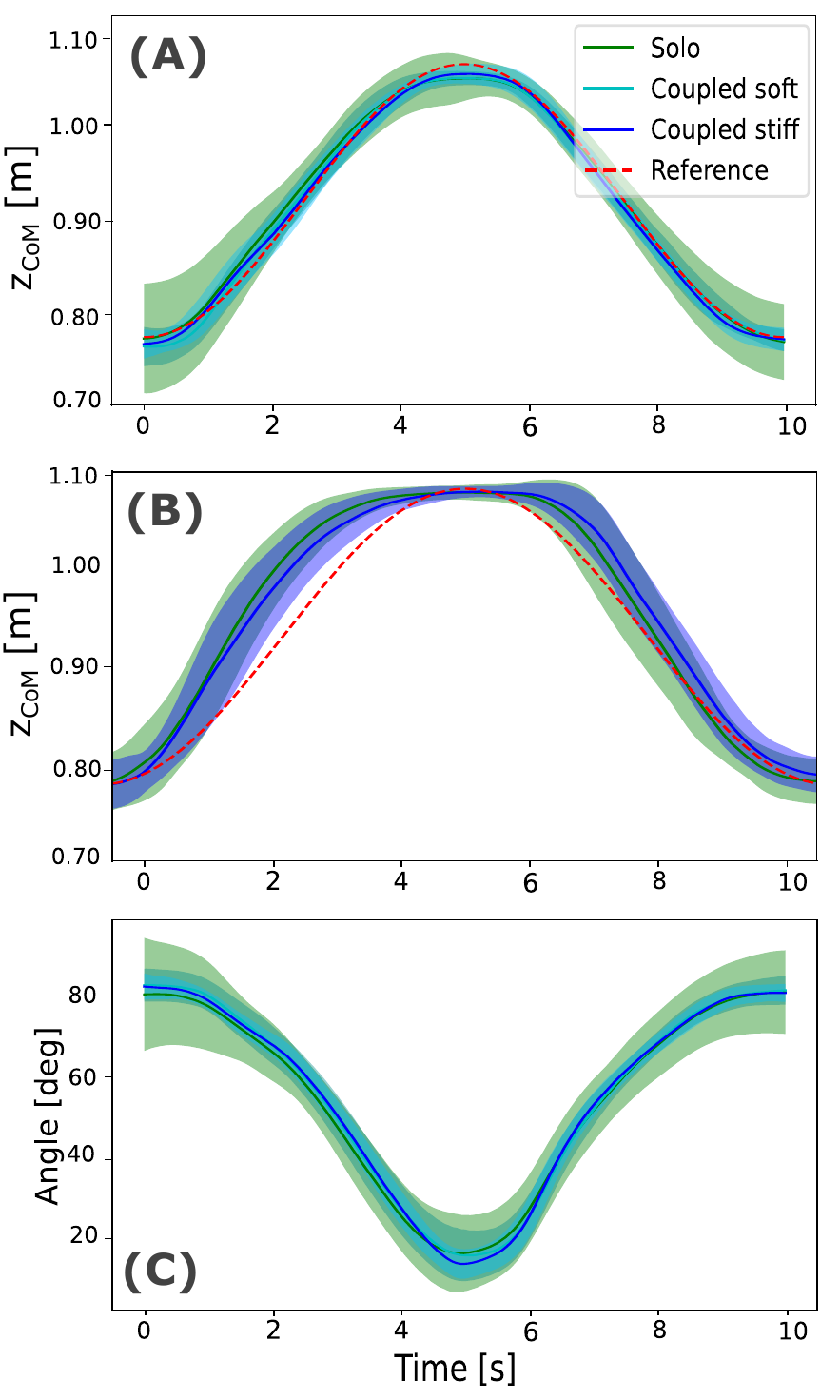}
  \caption{Distribution of Sit-to-Stand motion: (A) task space ($z_{CoM}$) movement distribution under different interaction conditions; (B)  task space  ($z_{CoM}$) movement distribution while both exoskeletons produce an upward assistance component; (C) joint space ($\mt{q}$) movement distribution of the right knee angle.}
  \label{fig:distribution}
\end{figure}

In these initial observations, we did not observe any particular differences between applying a virtual connection in the task space or joint space. This is likely due to the small variability with which StS movements are performed at knee and hip levels. Fig. \ref{fig:distribution}C shows the limited amount of movement in the null space for a single joint (right knee) both during the solo and dyad conditions. We expect that greater differences may be observed in users with differences in anatomy, strength and flexibility (e.g., therapist and patient). For instance, it is possible that interaction in the task space could reduce the quality of information the therapist can receive from the patient (i.e., lack of feedback in each joint)~\cite{brokaw2013comparison}. In future experiments, we intend to test higher stiffness values and compare users with different kinematic and dynamic properties. 


In Sec.\ref{sec:balancing}, we presented the balancing constraint we implemented  to compensate for the robot's movement limitations (i.e., non-actuated ankle joint).
In fact, ankle movements can lead to solutions that violate the constraint and thus to instability and unnatural movements for the user such as hyper-extension of the hip. The constraints on the vertical position and backpack angle are designed specifically to avoid these occurrences. For example, a forward movement of the ankle that violates the horizontal constraint generates solutions that lead the user to a safe position by lowering the CoM position and backpack angle. Backward movements, on the other hand, are less constrained to allow execution of the StS movement. Constraints on joint velocities and the backpack remain active, which limits dangerous movements. The results on the horizontal position of the CoM can be seen in Fig. \ref{fig:balancing_constraint}B,C.

The influence of vertical assistance from the robot was also evaluated in this study. Its contribution in guiding vertical movement and aiding descent was assessed through subjective feedback and comparison of CoM trajectories. Users reported difficulty in distinguishing the contribution of the assistance from that of the virtual connection. This is evident when comparing Fig. \ref{fig:distribution}A and Fig. \ref{fig:distribution}B: with assistance, 
in both solo and dyadic conditions, the users experienced a vertical force exerted by the exoskeleton, prompting them to adapt their approach. They responded by quickening their upward movements and postponing their downward movements. Based on these results, we believe that assistance from the robot is critical to limit the onset of fatigue, but it is better suited for discrete tracking tasks (Fig.\ref{fig:trials}C) in which users have more degree of freedom on the choice of the trajectory to execute.



\begin{figure}[t]
    \centering
    \vspace{0.2cm}
    \includegraphics[width=.9\linewidth]{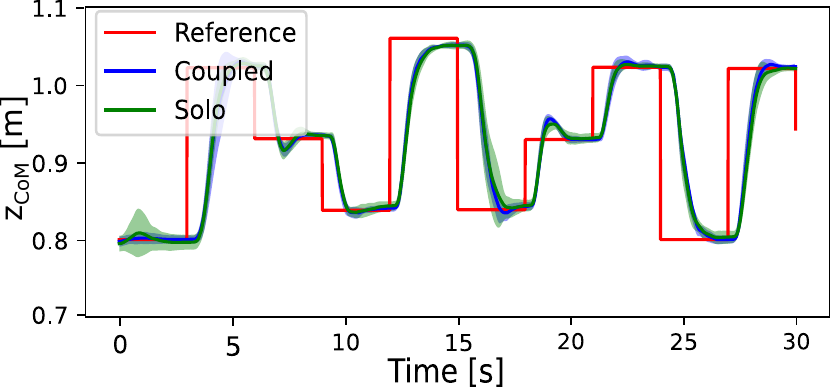}
    \caption{Example of a StS tracking task with discrete targets. }
    \label{fig:dicreteTask}
\end{figure}

In this study we present three different types of tracking tasks for a StS exercise (Fig. \ref{fig:trials}). Each has advantages: sinusoidal functions (Fig. \ref{fig:trials}A) were reported to be less-fatiguing because they allow users to pause in comfort positions (i.e., standing or sitting). However, the predictive nature of a single frequency sine function is likely too simple to analyze motor learning. Complex sinusoidal functions (Fig. \ref{fig:trials}B), on the other hand, promote variability of movement, which is fundamental to rehabilitative exercises. The drawback of multi-sine functions is that they are particularly fatiguing, requiring users to slow down and hold flexed postures to follow the speed of the reference. The discrete sinusoidal trajectory (Fig. \ref{fig:trials}C and Fig. \ref{fig:dicreteTask}) reduces these limitations, but is more difficult to evaluate from a performance standpoint for the variability with which users can choose to change their approach trajectory. Based on these results, we intend to perform additional pilot testing with more users to construct a rehabilitation task that appropriately challenges users while limiting fatigue over many repetitions. 

\section{Conclusion}

In this paper, we presented a framework for rehabilitation exercises specific to Sit-to-Stand movements based on the concept of exoskeleton mediated interaction. To share the task between the two participants, we proposed three types of tracking tasks. These tasks are aimed at promoting repeatability and at the same time movement variability, which are essential to activate motor learning in patients. We then extended the controller for exoskeleton mediated haptic interaction by including scalable and task-specific Sit-to-Stand assistance and support to ensure balance during movement execution. 
In the future, we intend to test this approach first on a healthy population to validate the learning ability of the Sit-to-Stand task and then on a population of therapists and patients to test its usability and contribution in the context of physical rehabilitation. 
Furthermore, we intend to include Machine Learning algorithms in this framework to compensate for any inconsistencies between movements or delays due to networking problems (as can occur in remote rehabilitation).

\section*{Acknowledgments}

This work was supported by the National Science Foundation~/~National Robotics Initiative (Grant No: 2024488).  We would like to thank Tim Haswell for his technical support on the hardware improvements of the ExoMotus-X2 exoskeleton.

\bibliographystyle{IEEEtran}
\bibliography{references,ref_rehab}

\begin{thebibliography}{10}
\providecommand{\url}[1]{#1}
\csname url@samestyle\endcsname
\providecommand{\newblock}{\relax}
\providecommand{\bibinfo}[2]{#2}
\providecommand{\BIBentrySTDinterwordspacing}{\spaceskip=0pt\relax}
\providecommand{\BIBentryALTinterwordstretchfactor}{4}
\providecommand{\BIBentryALTinterwordspacing}{\spaceskip=\fontdimen2\font plus
\BIBentryALTinterwordstretchfactor\fontdimen3\font minus
  \fontdimen4\font\relax}
\providecommand{\BIBforeignlanguage}[2]{{%
\expandafter\ifx\csname l@#1\endcsname\relax
\typeout{** WARNING: IEEEtran.bst: No hyphenation pattern has been}%
\typeout{** loaded for the language `#1'. Using the pattern for}%
\typeout{** the default language instead.}%
\else
\language=\csname l@#1\endcsname
\fi
#2}}
\providecommand{\BIBdecl}{\relax}
\BIBdecl

\bibitem{boukadida2015determinants}
A.~Boukadida, F.~Piotte, P.~Dehail, and S.~Nadeau, ``Determinants of
  sit-to-stand tasks in individuals with hemiparesis post stroke: a review,''
  \emph{Annals of physical and rehabilitation medicine}, vol.~58, no.~3, pp.
  167--172, 2015.

\bibitem{cheng1998sit}
P.-T. Cheng, M.-Y. Liaw, M.-K. Wong, F.-T. Tang, M.-Y. Lee, and P.-S. Lin,
  ``The sit-to-stand movement in stroke patients and its correlation with
  falling,'' \emph{Archives of physical medicine and rehabilitation}, vol.~79,
  no.~9, pp. 1043--1046, 1998.

\bibitem{mccrory2014work}
B.~McCrory, J.~M. Burnfield, A.~R. Darragh, J.~L. Meza, S.~L. Irons,
  P.~Chernyavskiy, A.~M. Link, and G.~Brusola, ``Work injuries among therapists
  in physical rehabilitation,'' in \emph{Proceedings of the Human Factors and
  Ergonomics Society Annual Meeting}, vol.~58, no.~1.\hskip 1em plus 0.5em
  minus 0.4em\relax SAGE Publications Sage CA: Los Angeles, CA, 2014, pp.
  1072--1076.

\bibitem{burnfield2013comparative}
J.~M. Burnfield, B.~McCrory, Y.~Shu, T.~W. Buster, A.~P. Taylor, and A.~J.
  Goldman, ``Comparative kinematic and electromyographic assessment of
  clinician-and device-assisted sit-to-stand transfers in patients with
  stroke,'' \emph{Physical Therapy}, vol.~93, no.~10, pp. 1331--1341, 2013.

\bibitem{fray2019evaluation}
M.~Fray, S.~Hignett, A.~Reece, S.~Ali, and L.~Ingram, ``An evaluation of sit to
  stand devices for use in rehabilitation,'' in \emph{Proceedings of the 20th
  Congress of the International Ergonomics Association (IEA 2018) Volume VII:
  Ergonomics in Design, Design for All, Activity Theories for Work Analysis and
  Design, Affective Design 20}.\hskip 1em plus 0.5em minus 0.4em\relax
  Springer, 2019, pp. 774--783.

\bibitem{Kucuktabak2021}
\BIBentryALTinterwordspacing
E.~B. K{\"u}{\c{c}}{\"u}ktabak, S.~J. Kim, Y.~Wen, K.~Lynch, and J.~L. Pons,
  ``Human-machine-human interaction in motor control and rehabilitation: a
  review,'' \emph{Journal of NeuroEngineering and Rehabilitation}, 2021.
  [Online]. Available: \url{https://doi.org/10.1186/s12984-021-00974-5}
\BIBentrySTDinterwordspacing

\bibitem{camardella2021gait}
C.~Camardella, F.~Porcini, A.~Filippeschi, S.~Marcheschi, M.~Solazzi, and
  A.~Frisoli, ``Gait phases blended control for enhancing transparency on
  lower-limb exoskeletons,'' \emph{IEEE Robotics and Automation Letters},
  vol.~6, no.~3, pp. 5453--5460, 2021.

\bibitem{sharifi2021adaptive}
M.~Sharifi, J.~K. Mehr, V.~K. Mushahwar, and M.~Tavakoli, ``Adaptive cpg-based
  gait planning with learning-based torque estimation and control for
  exoskeletons,'' \emph{IEEE Robotics and Automation Letters}, vol.~6, no.~4,
  pp. 8261--8268, 2021.

\bibitem{andrade2021trajectory}
R.~M. Andrade, S.~Sapienza, E.~E. Fabara, and P.~Bonato, ``Trajectory tracking
  impedance controller in 6-dof lower-limb exoskeleton for over-ground walking
  training: Preliminary results,'' in \emph{2021 international symposium on
  medical robotics (ISMR)}.\hskip 1em plus 0.5em minus 0.4em\relax IEEE, 2021,
  pp. 1--6.

\bibitem{huo2021impedance}
W.~Huo, H.~Moon, M.~A. Alouane, V.~Bonnet, J.~Huang, Y.~Amirat,
  R.~Vaidyanathan, and S.~Mohammed, ``Impedance modulation control of a
  lower-limb exoskeleton to assist sit-to-stand movements,'' \emph{IEEE
  Transactions on Robotics}, vol.~38, no.~2, pp. 1230--1249, 2021.

\bibitem{kuccuktabak2023haptic}
E.~B. K{\"u}{\c{c}}{\"u}ktabak, Y.~Wen, S.~J. Kim, M.~Short, D.~Ludvig,
  L.~Hargrove, E.~Perreault, K.~Lynch, and J.~Pons, ``Haptic transparency and
  interaction force control for a lower-limb exoskeleton,'' \emph{arXiv
  preprint arXiv:2301.06244}, 2023.

\bibitem{kuccuktabak2023virtual}
E.~B. K{\"u}{\c{c}}{\"u}ktabak, Y.~Wen, M.~Short, E.~Demirba{\c{s}}, K.~Lynch,
  and J.~Pons, ``Virtual physical coupling of two lower-limb exoskeletons,''
  \emph{arXiv preprint arXiv:2307.06479}, 2023.

\bibitem{short2023haptic}
\BIBentryALTinterwordspacing
M.~R. Short, D.~Ludvig, E.~B. Küçüktabak, Y.~Wen, L.~Vianello, E.~J.
  Perreault, L.~Hargrove, K.~Lynch, and J.~Pons, ``{Haptic Human-Human
  Interaction During an Ankle Tracking Task: Effects of Virtual Connection
  Stiffness},'' 5 2023. [Online]. Available:
  \url{https://www.techrxiv.org/articles/preprint/Haptic_Human-Human_Interaction_During_an_Ankle_Tracking_Task_Effects_of_Virtual_Connection_Stiffness/23234069}
\BIBentrySTDinterwordspacing

\bibitem{kim2022effect}
S.~J. Kim, Y.~Wen, D.~Ludvig, E.~B. K{\"u}{\c{c}}{\"u}ktabak, M.~R. Short,
  K.~Lynch, L.~Hargrove, E.~J. Perreault, and J.~L. Pons, ``Effect of dyadic
  haptic collaboration on ankle motor learning and task performance,''
  \emph{IEEE Transactions on Neural Systems and Rehabilitation Engineering},
  2022.

\bibitem{medrano2023real}
R.~L. Medrano, G.~C. Thomas, C.~G. Keais, E.~J. Rouse, and R.~D. Gregg,
  ``Real-time gait phase and task estimation for controlling a powered ankle
  exoskeleton on extremely uneven terrain,'' \emph{IEEE Transactions on
  Robotics}, 2023.

\bibitem{osqp}
\BIBentryALTinterwordspacing
B.~Stellato, G.~Banjac, P.~Goulart, A.~Bemporad, and S.~Boyd, ``{OSQP}: an
  operator splitting solver for quadratic programs,'' \emph{Mathematical
  Programming Computation}, vol.~12, no.~4, pp. 637--672, 2020. [Online].
  Available: \url{https://doi.org/10.1007/s12532-020-00179-2}
\BIBentrySTDinterwordspacing

\bibitem{hamza2020balance}
M.~F. Hamza, R.~A.~R. Ghazilla, B.~B. Muhammad, and H.~J. Yap, ``Balance and
  stability issues in lower extremity exoskeletons: A systematic review,''
  \emph{Biocybernetics and Biomedical Engineering}, vol.~40, no.~4, pp.
  1666--1679, 2020.

\bibitem{farkhatdinov2019assisting}
I.~Farkhatdinov, J.~Ebert, G.~Van~Oort, M.~Vlutters, E.~Van~Asseldonk, and
  E.~Burdet, ``Assisting human balance in standing with a robotic
  exoskeleton,'' \emph{IEEE Robotics and automation letters}, vol.~4, no.~2,
  pp. 414--421, 2019.

\bibitem{ramos2019dynamic}
J.~Ramos and S.~Kim, ``Dynamic locomotion synchronization of bipedal robot and
  human operator via bilateral feedback teleoperation,'' \emph{Science
  Robotics}, vol.~4, no.~35, p. eaav4282, 2019.

\bibitem{penco2018robust}
L.~Penco, B.~Cl{\'e}ment, V.~Modugno, E.~M. Hoffman, G.~Nava, D.~Pucci, N.~G.
  Tsagarakis, J.-B. Mouret, and S.~Ivaldi, ``Robust real-time whole-body motion
  retargeting from human to humanoid,'' in \emph{2018 IEEE-RAS 18th
  International Conference on Humanoid Robots (Humanoids)}.\hskip 1em plus
  0.5em minus 0.4em\relax IEEE, 2018, pp. 425--432.

\bibitem{vianello2021human}
L.~Vianello, L.~Penco, W.~Gomes, Y.~You, S.~M. Anzalone, P.~Maurice, V.~Thomas,
  and S.~Ivaldi, ``Human-humanoid interaction and cooperation: a review,''
  \emph{Current Robotics Reports}, pp. 1--14, 2021.

\bibitem{harib2018feedback}
O.~Harib, A.~Hereid, A.~Agrawal, T.~Gurriet, S.~Finet, G.~Boeris, A.~Duburcq,
  M.~E. Mungai, M.~Masselin, A.~D. Ames \emph{et~al.}, ``Feedback control of an
  exoskeleton for paraplegics: Toward robustly stable, hands-free dynamic
  walking,'' \emph{IEEE Control Systems Magazine}, vol.~38, no.~6, pp. 61--87,
  2018.

\bibitem{xu2017adaptive}
F.~Xu, X.~Lin, H.~Cheng, R.~Huang, and Q.~Chen, ``Adaptive stair-ascending and
  stair-descending strategies for powered lower limb exoskeleton,'' in
  \emph{2017 IEEE International Conference on Mechatronics and Automation
  (ICMA)}.\hskip 1em plus 0.5em minus 0.4em\relax IEEE, 2017, pp. 1579--1584.

\bibitem{deng2018human}
C.~Deng and Z.~Li, ``Human-guided robotic exoskeleton cooperative walking for
  climbing stairs,'' in \emph{2018 3rd International Conference on Advanced
  Robotics and Mechatronics (ICARM)}.\hskip 1em plus 0.5em minus 0.4em\relax
  IEEE, 2018, pp. 60--65.

\bibitem{menga2018lower}
G.~Menga and M.~Ghirardi, ``Lower limb exoskeleton for rehabilitation with
  improved postural equilibrium,'' \emph{Robotics}, vol.~7, no.~2, p.~28, 2018.

\bibitem{rodriguez2022wearable}
G.~Rodriguez~Tapia, I.~Doumas, T.~Lejeune, and J.-G. Previnaire, ``Wearable
  powered exoskeletons for gait training in tetraplegia: a systematic review on
  feasibility, safety and potential health benefits,'' \emph{Acta Neurologica
  Belgica}, vol. 122, no.~5, pp. 1149--1162, 2022.

\bibitem{ishiguro2020bilateral}
Y.~Ishiguro, T.~Makabe, Y.~Nagamatsu, Y.~Kojio, K.~Kojima, F.~Sugai,
  Y.~Kakiuchi, K.~Okada, and M.~Inaba, ``Bilateral humanoid teleoperation
  system using whole-body exoskeleton cockpit tablis,'' \emph{IEEE Robotics and
  Automation Letters}, vol.~5, no.~4, pp. 6419--6426, 2020.

\bibitem{FongCanOpenDevelopment}
J.~Fong, E.~B. Küçüktabak, V.~Crocher, Y.~Tan, K.~Lynch, J.~Pons, and
  D.~Oetomo, ``{CANopen Robot Controller (CORC): An open software stack for
  human robot interaction development},'' \emph{The International Symposium on
  Wearable Robotics (WeRob2020)}, vol. 2020, 2020.

\bibitem{brokaw2013comparison}
E.~B. Brokaw, R.~J. Holley, and P.~S. Lum, ``Comparison of joint space and end
  point space robotic training modalities for rehabilitation of interjoint
  coordination in individuals with moderate to severe impairment from chronic
  stroke,'' \emph{IEEE Transactions on Neural Systems and Rehabilitation
  Engineering}, vol.~21, no.~5, pp. 787--795, 2013.

\end{thebibliography}

\end{document}